\pdfoutput=1

\documentclass[11pt]{article}

\usepackage[final]{acl}

\usepackage{times}
\usepackage{latexsym}

\usepackage[T1]{fontenc}

\usepackage[utf8]{inputenc}

\usepackage{microtype}

\usepackage{inconsolata}

\usepackage{graphicx}

\usepackage{booktabs}
\usepackage{amsmath}
\usepackage{amsfonts}
\usepackage{multirow}
\usepackage[export]{adjustbox}
\usepackage{txfonts}
\usepackage{enumitem}
\setlist[itemize]{align=parleft,left=0pt..1em}
\usepackage{array}
\usepackage{float}
\usepackage{graphicx}
\newcommand\blfootnote[1]{%
  \begingroup
  \renewcommand\thefootnote{}\footnote{#1}%
  \addtocounter{footnote}{-1}%
  \endgroup
}
\title{IFCap: Image-like Retrieval and Frequency-based Entity Filtering \\for Zero-shot Captioning}

\author{Soeun Lee$^*$\quad
        Si-Woo Kim$^*$\quad
        Taewhan Kim\quad
        Dong-Jin Kim$^{\dagger}$ \\
        Hanyang University, South Korea. \\
        {\footnotesize{\texttt{\{soeun, boreng0817, taewhan, djdkim\}@hanyang.ac.kr}}}}
\newcommand{\mff}[1]{\textbf{#1}}
\newcommand{\mss}[1]{\underline{#1}}
\newcommand{\model}{\text{IFCap}}

\newcommand{\Tref}[1]{Table~\ref{#1}}

\newcommand{\Fref}[1]{Fig.~\ref{#1}}

\newcommand{\Sref}[1]{Sec.~\ref{#1}}

\definecolor{ao(english)}{rgb}{0.0, 0.5, 0.0}

\begin{document}
\maketitle
\begin{abstract}
Recent advancements in image captioning have explored text-only training methods to overcome the limitations of paired image-text data. However, existing text-only training methods often overlook the modality gap between using text data during training and employing images during inference. To address this issue, we propose a novel approach called Image-like Retrieval, which aligns text features with visually relevant features to mitigate the modality gap. Our method further enhances the accuracy of generated captions by designing a Fusion Module that integrates retrieved captions with input features. Additionally, we introduce a Frequency-based Entity Filtering technique that significantly improves caption quality. We integrate these methods into a unified framework, which we refer to as $\model$ (\textbf{I}mage-like Retrieval and \textbf{F}requency-based Entity Filtering for Zero-shot \textbf{Cap}tioning). Through extensive experimentation, our straightforward yet powerful approach has demonstrated its efficacy, outperforming the state-of-the-art methods by a significant margin in both image captioning and video captioning compared to zero-shot captioning based on text-only training.\blfootnote{$^*$Equal contribution. $^{\dagger}$Corresponding author.}\footnote{Code: \url{https://github.com/boreng0817/IFCap}}
\end{abstract}

\section{Introduction}

\begin{figure*}
    \centering    \includegraphics[width=1.04\textwidth]{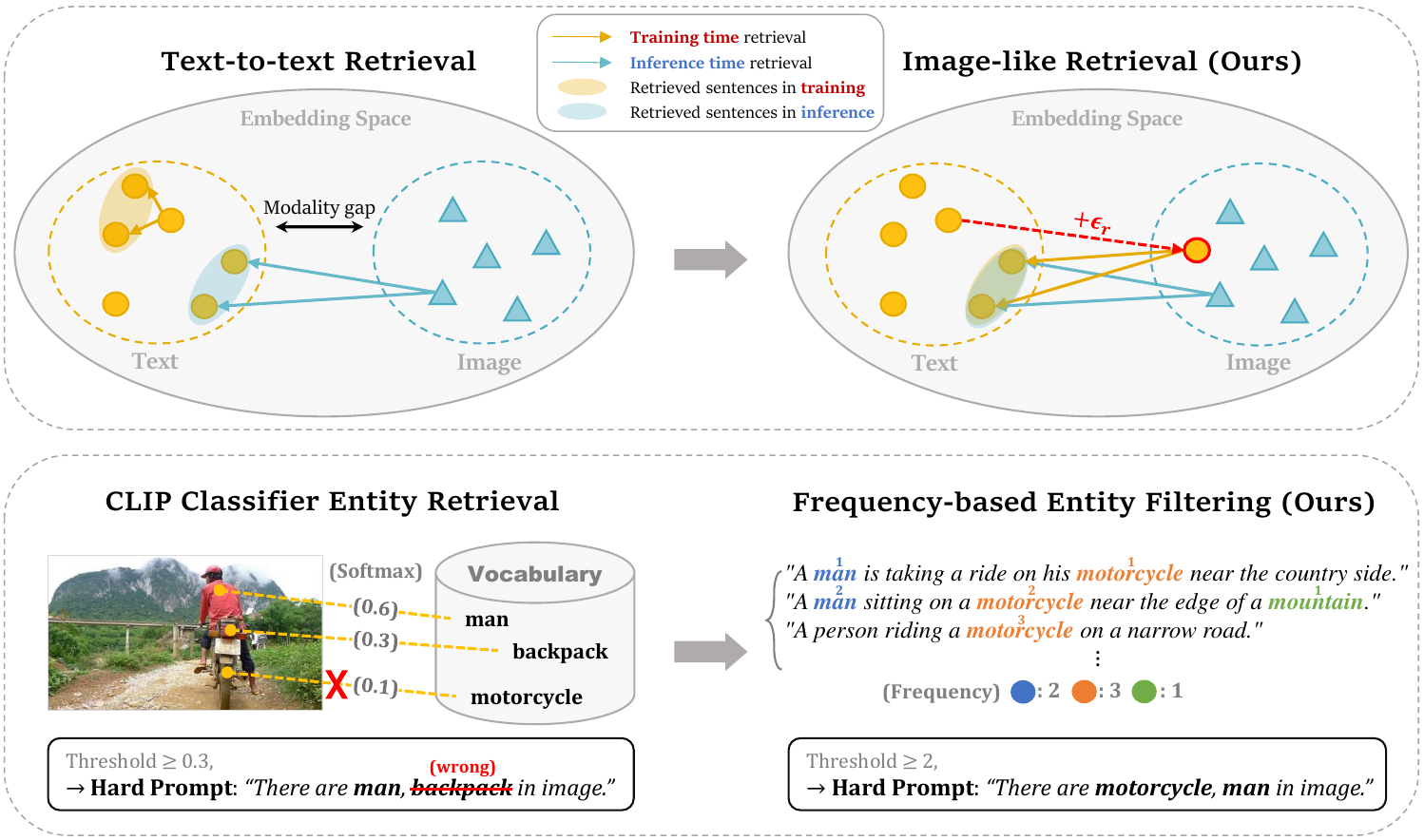}
    \caption{(Top) The previous text-to-text retrieval approach overlooks the modality gap, leading to different information use between training and inference. Our approach addresses this by aligning text features with the image embedding space during retrieval. (Bottom) The traditional CLIP classifier-based entity retrieval method struggles with entity detection as vocabulary size grows. Our approach detects frequently occurring words in retrieved captions, extracting entities more accurately without relying on a limited vocabulary.}
    \label{fig:teaser}
\end{figure*}

The task of image captioning generates appropriate textual descriptions for images by combining computer vision (CV) and natural language processing (NLP).  With the emergence of Large Language Models (LLMs) and Vision and Language Models (VLMs), various works have studied efficient training methods for image captioning~\cite{clipcap, ituning, smallcap}. These approaches develop effective captioning by using pre-trained models with few parameters or lightweight networks. However, these works rely on paired image-text data, which is costly~\cite{kim2019image,kim2024semi}. To overcome this, recent studies have explored text-only training methods for image captioning, aiming to solve the problem using only textual data \cite{capdec, decap, viecap, meacap, knight, syntic, icsd}.

Text-only training introduces a new direction in which models are trained solely using text data. Recent existing works have studied what to use as extra cues, such as extracted nouns~\cite{viecap}, generated synthetic images~\cite{syntic,icsd} for training, and extracted tags from object detectors~\cite{syntic}.
However, existing methods that rely on object information are sensitive to incorrect data, and utilizing large external models (e.g., stable diffusion~\citealp{stablediffusion} or object detectors~\citealp{detr}) incurs additional costs. Thus, we aim to address the problem by acquiring diverse information cost-effectively without additional models.

The retrieval task involves finding relevant information in a database for a given query. Initially rooted in NLP \cite{rag}, the field has expanded into CV and into multi-modal retrieval. 
Depending on the input data and database, various retrieval methods are possible, such as image-to-text \cite{smallcap} and text-to-text retrieval \cite{knight}.
In the existing text-only training study, there have been attempts to use the text-to-text retrieval method. However, existing works can't address the modality gap inherent in text-only training settings, where training is performed with text and inference with images. In addition, such works rely too much on retrieved captions without considering visual information. This modality gap and the use of a narrow scope of information may lead to performance degradation.

To verify this, we visualize the analysis result of the CLIP embedding feature of retrieved captions that the model uses in training via t-SNE in \Fref{fig:analysis1}.
The analysis is done on the COCO~\cite{mscoco} validation split, and the CLIP similarity-based KNN algorithm is used for retrieval.
In the figure, there is a large difference between the distribution of features used after image-to-text retrieval and text-to-text retrieval, which shows that a modality gap exists between image and text. 

To tackle this issue, we propose a novel approach called ``Image-like Retrieval,'' that addresses the modality gap between image and text data. We inject a noise into the CLIP text feature to act as a query in image feature distribution. 
Visualization results for this approach are shown in \Fref{fig:analysis1} right, demonstrating that our method exhibits a distribution highly similar to that of image-to-retrieval results and ground truth captions, unlike traditional text-to-text retrieval methods. Indeed, when our method is applied to the existing research \cite{knight}, performance improvements are observed, as shown in Table ~\ref{tb:knight_improvement}.

Prior research \cite{knight} relies solely on retrieved captions, which may include wrong information in the input caption, potentially leading to inaccurate outputs. To address this, we design a \emph{Fusion Module} that effectively integrates both the original input and additional representations.
Additionally, as shown by numerous studies \cite{viecap, smallcap}, prompts can clarify the information provided to the language model. We extract keywords from the input caption to construct a hard prompt, which is fed to the LLM, offering explicit guidance. This approach maximizes the utility of text data, guiding the model to generate accurate and relevant captions.

\begin{figure}[t]
    \centering
    \includegraphics[width=\linewidth]{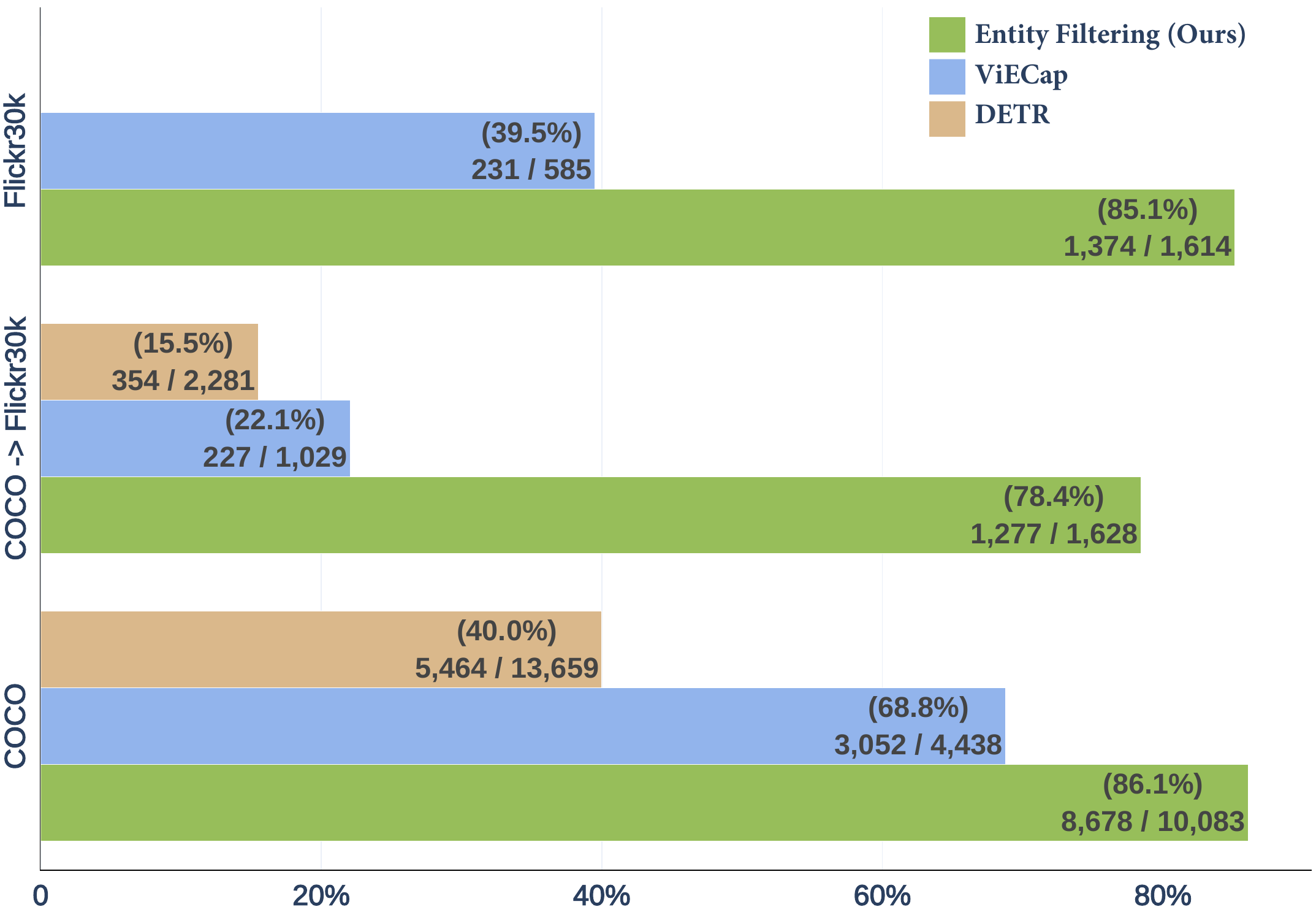}
    \caption{Precision of extracted entities in the COCO test set, total 5,000 images. If an extracted entity exists in the ground-truth caption, it counts as correct or else wrong. Three methods (Ours, ViECap\citeyear{viecap}, DETR\citeyear{detr}) are compared with three different settings. Our method is illustrated in \ref{3.3}, and ViECap uses CLIP based classifier with the source domain's vocabulary list. We follow the way SynTIC \cite{syntic} uses DETR and employ the COCO vocabulary list. Due to the inaccessible vocabulary list of Flickr30k, DETR can't be compared, and ViECap uses the VGOI~\cite{vgoi} vocabulary list in Flickr30k. Our method dominates the precision score and quantity of entities in every setting.}
    \vspace{-2mm}
    \label{fig:analysis2}
\end{figure}

Guiding caption decoder with extracted entities from an image helps the model generate an accurate description of the image. However, {we find that the previous works ~\cite{viecap, syntic} show low entity detection precision, especially when the vocabulary is large as shown in \Fref{fig:analysis2}.} 
{Therefore,}
we propose a Frequency-based Entity Filtering technique precisely utilizing entity information without relying on the vocabulary. During inference, we utilize retrieved sentences from images, 
parsing them into nouns and calculating their frequency. 
Then, we filter nouns with pre-defined thresholds and curate hard prompts for the text decoder. 
This simple method yields remarkable performance improvements.

In summary, our contributions are as follows:
\begin{itemize}[noitemsep,topsep=0pt]
\item We propose a novel approach, \emph{Image-like Retrieval}, which achieves effects similar to image-to-text retrieval in text-only training. Then, we introduce a \emph{Fusion Module} for interaction between existing and additional representations.
\item We propose an entity filtering technique in inference, \emph{Frequency-based Entity Filtering}, enhancing the language model by filtering frequently appearing entities in retrieved captions.
\item Extensive evaluations show $\model$ achieves state-of-the-art performance in various benchmarks, including video captioning.
\end{itemize}


\section{Related work}

\subsection{Text-only Captioning}
The advantage of CLIP \cite{clip} has been utilized in a variety of tasks, such as image captioning, image generation, and object detection.
In the realm of image captioning, text-only training research is emerging that uses only text data for learning without image data, taking advantage of the CLIP characteristic that image embeddings and text embeddings are learned to be close.
DeCap \cite{decap} trains a text decoder using only textual data and introduces a support memory mechanism to project input images into the text embedding space during inference, facilitating the generation of captions.
ViECap~\cite{viecap} recognizes the main entity of text data that comes as input and configures it as a prompt, allowing LLM to perform object-agnostic learning based on open vocabulary retrieval using CLIP.

\begin{figure}[t]
    \includegraphics[width=0.48\textwidth]{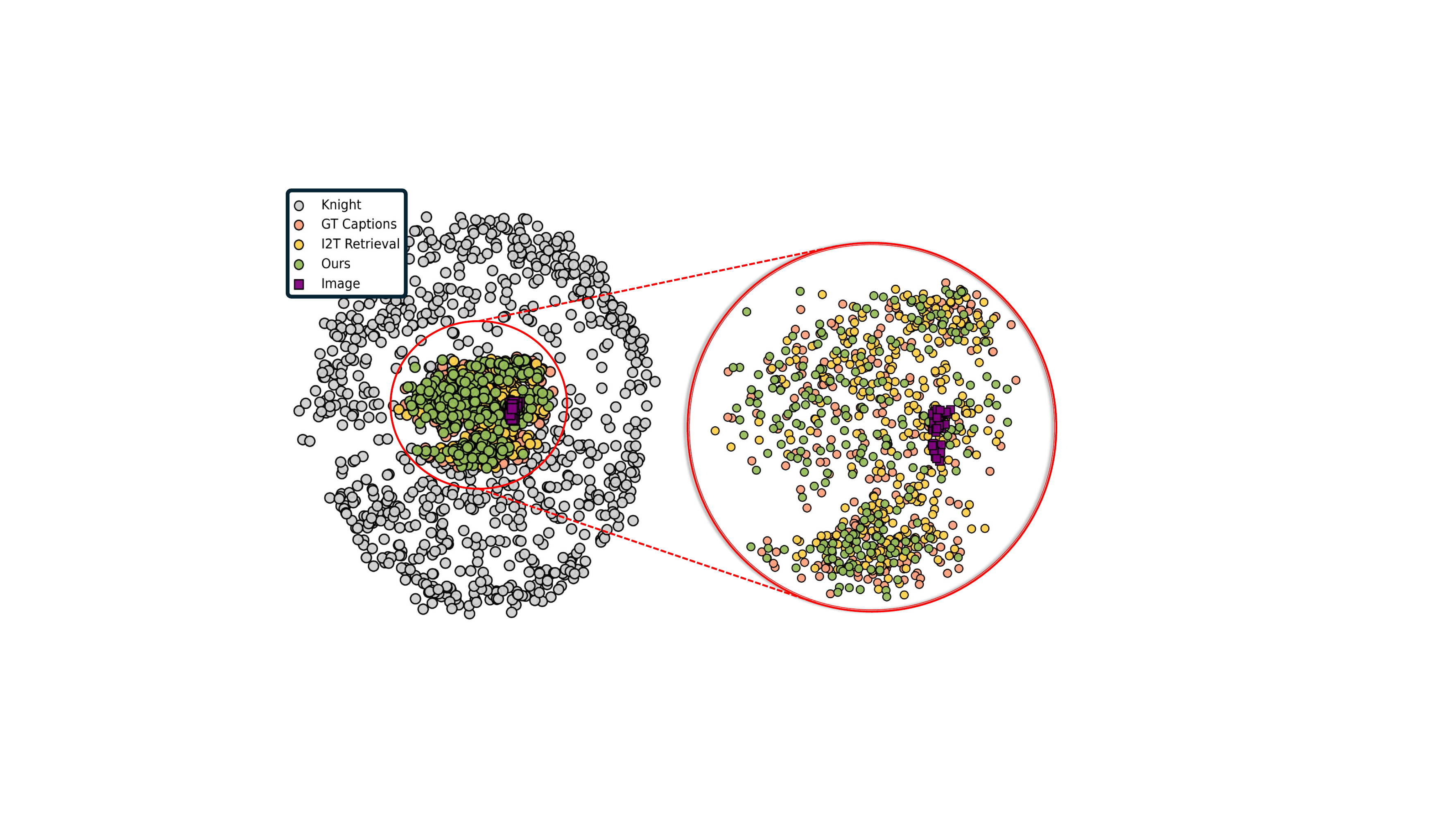}        
    \caption{The distribution of CLIP embedding features corresponding to images $\textcolor{violet}{\blacksquare}$, paired captions $\textcolor{red}{\medbullet}$, retrieved captions $\textcolor{yellow}{\medbullet}$ for a specific image, and the result of text-to-text retrieval  $\textcolor{gray}{\medbullet}$ and our Image-like Retrieval $\textcolor{ao(english)}{\medbullet}$.}
    \label{fig:analysis1}
    \vspace{-2mm}
\end{figure}

\begin{figure*}
    \centering
    \includegraphics[width=1\textwidth]{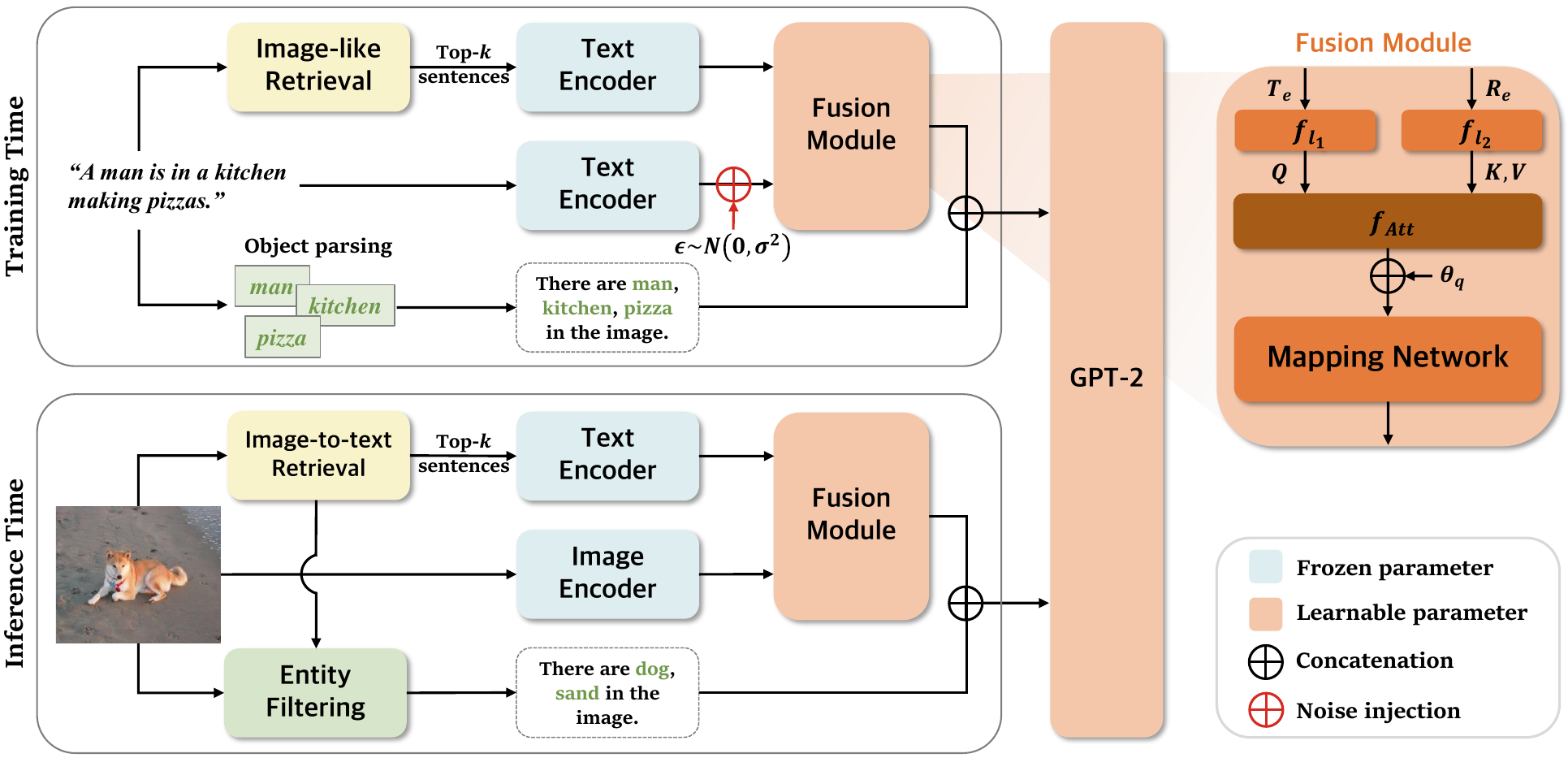}
    \caption{The overview of IFCap. During training, we extract nouns from the input text and retrieve $k$ similar sentences using our Image-like Retrieval method. Extracted nouns are incorporated into a prompt template to form a hard prompt. Both the input text and retrieved sentences are encoded using the text encoder. These embeddings interact and combine through our Fusion Module before being fed into the LLM for sentence generation. During inference, we retrieve $l$ sentences similar to the input image and construct a hard prompt by extracting entities via Frequency-based Entity Filtering from the retrieved sentences. The sentences are encoded using a text encoder, and the input image is encoded using an image encoder, followed by input into the Fusion Module. The subsequent process follows a procedure similar to the training phase.}
    \label{fig:model}
\end{figure*}

\subsection{Modality Gap}
Vision language models such as CLIP aim to embed images and text closely in a shared space. However, it has been shown that these embeddings are located in two separate regions, with a significant gap between the modalities \cite{mind}. This modality gap hinders the interaction between vision and text modalities and limits the quality of generated captions. Among the notable approaches addressing this issue, CapDec \cite{capdec} assumes that the image embeddings paired with text embeddings are located within a small radius around the text embeddings and mitigates the gap with noise injection. CLOSE \cite{close} highlights the low cosine similarity between images and their paired texts and uses a hyper-parameter-scaled noise injection technique to bridge the gap.

We focus on the modality gap for retrieval from a new perspective. Our goal is to perform text retrieval similar to image-to-text retrieval, considering the modality gap. The distinction from existing methods can be observed in \Fref{fig:analysis1} left.

\subsection{Retrieval Augmented Generation}
Retrieval has been used in diverse ways in NLP. Image captioning also benefits from retrieval modules by incorporating novel objects and new information into captions, allowing access to new domains without additional training. Retrieval is applied in various ways in image captioning models. For instance, Smallcap~\cite{smallcap} retrieves captions relevant to the input image and uses them as instructions for the text decoder. In text-only image captioning, ViECap~\cite{viecap} retrieves novel objects from the input image and uses them as prompts, while Knight~\cite{knight} uses retrieved captions as text features.


Most retrieval methods are based on image-to-text retrieval, but text-only captioning performs text-to-text retrieval. However, during inference, the modality gap caused by the input image leads to poor performance. Our method carefully addresses this issue to improve performance by considering the gap between image and text.
\section{Methods}

We propose a new text-only image captioning model, \model, which is illustrated in ~\Fref{fig:model}. During training, the model only utilizes text data, as is standard for text-only training models. First, we embed the input text using a text encoder. The text embeddings are then fed into a mapping network to close the gap between different modalities. Finally, the processed embeddings go through a caption decoder to generate the output caption.


Our $\model$ utilizes a simple yet powerful retrieval mechanism and addresses the modality gap between image and text with Image-like Retrieval (Section \ref{3.1}). After performing Image-like Retrieval, we employ a Fusion Module (Section \ref{3.2}) to merge input embeddings with the retrieved features. During inference, we use the retrieved captions from the image to find accurate and detailed entities with  Frequency-based Entity Filtering (Section \ref{3.3}).

\subsection{Image-like Retrieval (ILR)} \label{3.1}
While text-to-text retrieval can be effectively performed during training, it is likely to suffer from performance degradation during inference when an image is provided as input due to the modality gap. Therefore, Image-like Retrieval (ILR) aims to perform text-to-text retrieval in a manner that resembles image-to-text retrieval outcomes, given text input. For this, we propose an approach that inserts noise into the feature space of the input text, bringing it closer to the image feature space. The augmentation process is as follows:

First, we utilize the CLIP to embed the input text $t_i$  and the text corpus $\boldsymbol{\mathcal{T}}=\{t_i\}_{i=1}^{N_c}$ with a text encoder $\mathcal{E}_T$. 
Then, we introduce noise $\epsilon_r \sim N(0, \sigma_r^2)$ into the embedding of input text $T_i$, aiming to adjust the text features to align more closely with the image feature space:\
\begin{gather}
    T_i=\mathcal{E}_T(t_i), ~~T_i^\epsilon = T_i + \epsilon_r.
\end{gather}
Next, the retrieval step is performed using the noise-injected input text ${T_i^\epsilon}$. To identify the descriptions most relevant to ${T_i^\epsilon}$, the top-$k$ descriptions are retrieved by calculating the cosine similarity between ${T_i^\epsilon}$ and all sentence embeddings in the text corpus. This process closely follows previous methods in image-to-text retrieval \cite{smallcap}, with the distinction that we perform retrieval based on ${T_i^\epsilon}$ instead of images.

By utilizing this approach during training, we can enhance the ability of a model to provide image-like information even in a text-only training setting, thereby narrowing the modality gap and improving performance.

\subsection{Fusion Module (FM)} \label{3.2}

In text-only image captioning, choosing which additional information to inject into the model and dealing with new representations with given data appropriately are important issues. To handle this problem, we use the attention mechanism~\cite{transformer} to fuse input text features and retrieved captions features to extract their meaningful interaction. The attention mechanism emphasizes certain important features, and due to its effectiveness, it has been widely utilized in the field of captioning \cite{show}.  

We first encode input text and retrieved captions using CLIP~\cite{clip} text encoder, then inject a Gaussian noise $\epsilon\sim N(0,\sigma^2)$ to input text feature for relieving the modality gap between image and text. Then we adjust the dimension of the input text feature and retrieved captions feature to the embedding space of caption decoder with linear layers $f_{l_1}$ and $f_{l_2}$ respectively, and apply cross-attention $\boldsymbol{f}_{Att}$ with $T_e$ as query and $R_e$ as key, then create fusion representation $F_e$ containing input text and retrieved captions. Finally, $F_e$ is fed into a trainable Mapping Network, which encodes the overall contents of the given input. We can summarize this process with equations.
\begin{align}
    T_e &= T_i+\epsilon, ~~~ R_e=\mathcal{E}_T(\text{ILR}(T_i)), \\
    F_e &= \boldsymbol{f}_{Att}(f_{l_1}(T_e), f_{l_2}(R_e)), \\
    \boldsymbol{F} &= \text{Map}(F_e;\theta_q).
\end{align}
The noun implies intuitive and explicit information about objects in the image. For employing the property of nouns, we extract entities in each training text corpus and input images. We build a hard prompt $h$ with extracted entities $E = \{e_1, e_2, ..., e_n\}$ to make the model aware of existing entities in the image. With retrieved captions and hard prompts with entities, the model can learn the ability to generate proper captions without images. We use auto-regressive loss to optimize our projector and caption decoder. (Details about the Fusion Module are in ~\Sref{4.1}).
\begin{equation}
        L_\theta=-{\frac{1}{N}}\sum_{i=1}^N{\log({y_i}|\boldsymbol{F};\boldsymbol{h};y_{<i};\theta}).
\end{equation}

\subsection{Frequency-based Entity Filtering (EF)} \label{3.3}
After retrieving $l$ captions from an image, we use grammar parser tools (e.g., NLTK~\citealp{nltk}) to extract nouns from the retrieved sentences and calculate the frequency of these extracted nouns as $F=[f_1, f_2,...,f_n]$. We then select nouns that have a frequency larger than a predefined threshold and place them into a hard prompt.

$\textbf{Heuristic threshold}$: Since frequency is discrete, we can manually find the best threshold by conducting experiments with every possible threshold. This allows us to determine the global optimal threshold.

\begin{table*}
\tabcolsep=5pt
\centering
\resizebox{\linewidth}{!}{
\begin{tabular}{l|c|c|cccc|cccc}
\toprule[2pt]
\multirow{2}{*}{\textbf{Method}} & \textbf{Image} & \textbf{Text} & \multicolumn{4}{c|}{\textbf{COCO}} & \multicolumn{4}{c}{\textbf{Flickr30k}} \\
& \textbf{Encoder} & \textbf{Decoder} & \multicolumn{1}{c}{B@4} & \multicolumn{1}{c}{M} & \multicolumn{1}{c}{C} & \multicolumn{1}{c|}{S} & \multicolumn{1}{c}{B@4} & \multicolumn{1}{c}{M} & \multicolumn{1}{c}{C} & \multicolumn{1}{c}{S} \\
\midrule
CapDec \citeyearpar{capdec}                        & RN50x4   & GPT-2$_\text{Large}$                &      26.4  &      25.1  &       91.8  &      11.9  &      17.7  &      20.0 &       39.1 &       9.9  \\
DeCap  \citeyearpar{decap}                         & ViT-B/32 & Transformer$_\text{Base}$           &      24.7  &      25.0  &       91.2  &      18.7  &      21.2  &      21.8 &       56.7 &       15.2 \\
CLOSE  \citeyearpar{close}                         & ViT-L/14 & T5$_\text{base}$                    &      -     &      -     &       95.3  &      -     &      -     &      -    &       -    &       -    \\
ViECap \citeyearpar{viecap}                        & ViT-B/32 & GPT-2$_\text{Base}$                 &      27.2  &      24.8  &       92.9  &      18.2  &      21.4  &      20.1 &       47.9 &       13.6 \\
MeaCap$_\text{InvLM}$ \citeyearpar{meacap}         & ViT-B/32 & GPT-2$_\text{Base}$                 &      27.2  &      25.3  &       95.4  &      19.0  &      22.3  &      22.3 & \mss{59.4} &       15.6 \\
Knight \citeyearpar{knight}                        & RN50x64  & GPT-2$_\text{Large}$                &      27.8  & \mss{26.4} &       98.9  & \mss{19.6} &      22.6  & \mff{24.0}&       56.3 &       16.3 \\
ICSD$^\spadesuit$ \citeyearpar{icsd}               & ViT-B/32 & BERT$_\text{Base}$                  & \mss{29.9} &      25.4  &       96.6  &      -     & \mff{25.2} &      20.6 &       54.3 &       -    \\
SynTIC$^{\spadesuit\dagger}$ \citeyearpar{syntic}  & ViT-B/32 & Transformer$_\text{H=4}^\text{L=4}$ & \mss{29.9} &      25.8  & \mss{101.1} &      19.3  &      22.3  &      22.4 &       56.6 & \mss{16.6} \\
\midrule
\model                                             & ViT-B/32 & GPT-2$_\text{Base}$                 & \mff{30.8} & \mff{26.7} & \mff{108.0} & \mff{20.3} & \mss{23.5} & \mss{23.0}& \mff{64.4} & \mff{17.0} \\
\bottomrule[2pt]
\end{tabular}
}\caption{Result on the In-domain captioning including COCO test split and Flickr30k test split. Every result is copied from the original papers. $\spadesuit$: Utilizes text-to-image generation model in the training time, $\dagger$: Utilizes object detector during the training and inference time.  $\model$ achieves state-of-the-art in most metrics. The best number overall is in \mff{bold} and the second best in \mss{underline}.}\label{tb:indomain}
\end{table*} 
\begin{table}[ht]
\tabcolsep=4pt
\resizebox{\columnwidth}{!}{
\begin{tabular}{l|cccc|cccc}
\toprule[2pt]
\multirow{2}{*}{\textbf{Method}} & \multicolumn{4}{c|}{\textbf{COCO $\Longrightarrow$ Flickr}} & \multicolumn{4}{c}{\textbf{Flickr $\Longrightarrow$ COCO}} \\
& \multicolumn{1}{c}{B@4} & \multicolumn{1}{c}{M} & \multicolumn{1}{c}{C} & \multicolumn{1}{c|}{S} & \multicolumn{1}{c}{B@4} & \multicolumn{1}{c}{M} & \multicolumn{1}{c}{C} & \multicolumn{1}{c}{S} \\ 
\midrule
DeCap \citeyearpar{decap}    &      16.3  &      17.9  &      35.7  &      11.1  &      12.1  &      18.0  &      44.4  &      10.9  \\
ViECap \citeyearpar{viecap}  &      17.4  &      18.0  &      38.4  &      11.2  &      12.6  &      19.3  &      54.2  &      12.5  \\
Knight \citeyearpar{knight}  & \mss{21.1} & \mff{22.0} & \mss{48.9} & \mss{14.2} & \mss{19.0} & \mss{22.8} & \mss{64.4} & \mss{15.1} \\
SynTIC \citeyearpar{syntic}  &      17.9  &      18.6  &      38.4  &      11.9  &      14.6  &      19.4  &      47.0  &      11.9  \\
SynTIC-$TT$                  &      19.4  &      20.2  &      43.2  &      13.9  & \mff{20.6} &      21.3  & \mss{64.4} &      14.3  \\
\midrule
\model$^\star$               &     17.8   &       19.4 &       47.5 &       12.7 &       14.7 &       20.4 &       60.7 &      13.6  \\
\model-$TT$          & \mff{21.2} & \mss{21.8} & \mff{59.2} & \mff{15.6} & \mss{19.0} & \mff{23.0} & \mff{76.3} & \mff{17.3} \\
\bottomrule[2pt]
\end{tabular}
}\caption{Results on the Cross-domain captioning. $-TT$: models can access to target domain's corpus during inference time. $\star$: without Entity Filtering module in the inference time. $\model$ achieves state-of-the-art in most metrics.}\label{tb:crossdomain}
\end{table}


$\textbf{Adaptive threshold}$: We can use a heuristic threshold, but these thresholds are often unsuitable for different environments, and performing extensive experiments incurs unnecessary costs. Instead, we can estimate the common distribution of noun frequencies as certain probability distributions. We can assume frequencies follow $N(\mu_F, \sigma_F^2)$.
\begin{equation}
    \tau_\text{adap} = \mu_F + \sigma_F.
\end{equation}
Any nouns with a frequency larger than $\tau_\text{adap}$, which places them in the upper 15$\%$, can be considered outliers. Using this adaptive threshold, we can implement a flexible threshold that fits various settings. However, it does not guarantee global optima, leading to a trade-off relationship between heuristic thresholds and adaptive thresholds.
\section{Experiments}

\subsection{Implementation Details}\label{4.1}

While verifying the state-of-the-art performance of our model, we use CLIP (ViT-B/32) as the image encoder and GPT2$_\text{base}$~\cite{gpt2} as the text decoder. Parameters in the image encoder are frozen during training, and the text decoder and Fusion Module are trained. We train a total of 5 epochs, learning rate of $2\times10^{-5}$, use {scheduler} for learning rate scheduler, AdamW optimizer~\cite{adamw}, and set batch size 80. We use a single NVIDIA RTX4090 with 24GB VRAM; it takes about an hour and uses 12GB of VRAM during training. 

\textbf{Image-like Retrieval}: We first discover adequate $\sigma_r$ for Image-like Retrieval. Based on our experiment (\Fref{fig:analysis3}), we choose $\sigma_r$ as 0.04 in most cases. We retrieve $k$ sentences with noise-injected input text feature $T_e$. 

\textbf{Fusion Module}: We project $T_e\in\mathbb{R}^d$ and $R_e\in\mathbb{R}^{d\times k}$ with $f_{l_1}$, $f_{l_2}$ into $\mathbb{R}^{d_{gpt}}$, $\mathbb{R}^{d_{gpt}\times k}$ respectively where $d$ is the CLIP dimension and $d_{gpt}$ is the dimension of GPT-2 embedding space. We use projected $T_e$ as query and $R_e$ as key in $\boldsymbol{f}_{Att}$ layer. Finally, $F_e$ and $\theta_q$ are concatenated and fed into the Mapping Network, which consists of 8 layered transformers~\cite {transformer}. 

\noindent\textbf{Frequency-based Entity Filtering}: From the input image, we retrieve $l$ sentences and extracted nouns to obtain frequency $F$. With the predefined threshold, we filter entities and build hard prompt $\boldsymbol{h}$, providing more accurate and diverse entities to the caption decoder.

\noindent\textbf{Datasets, metrics} We evaluate our model in human-annotated datasets. For in-domain generalization, we test our model on MS-COCO~\cite{mscoco}, Flickr30k~\cite{flickr30k}, and utilize Karpathy split~\cite{Karpathy}. Also, to check the model's performance in the unseen scenarios, we use the NoCaps~\cite{nocaps} validation set. For metrics, we use common image captioning metrics CIDEr~\cite{cider}, SPICE~\cite{spice}, BLEU@$n$~\cite{bleu}, and METEOR~\cite{meteor}. More details about datasets and metrics are included in the appendix (\Sref{appendixD}).

\begin{table}[t]
\tabcolsep=3pt
\resizebox{\columnwidth}{!}{
\begin{tabular}{l|cc|cc|cc|cc}
\toprule[2pt]
\multirow{3}{*}{\textbf{Method}} & \multicolumn{8}{c}{\textbf{COCO $\Longrightarrow$ NoCaps Val}} \\
& \multicolumn{2}{c|}{In} & \multicolumn{2}{c|}{Near} & \multicolumn{2}{c|}{Out} & \multicolumn{2}{c}{Entire} \\
& \multicolumn{1}{c}{C} & \multicolumn{1}{c|}{S} & \multicolumn{1}{c}{C} & \multicolumn{1}{c|}{S} & \multicolumn{1}{c}{C} & \multicolumn{1}{c|}{S} & \multicolumn{1}{c}{C} & \multicolumn{1}{c}{S} \\ 
\midrule
DeCap  \citeyearpar{decap}   &      65.2 &         - &      47.8 &         - &      25.8 &        - &      45.9 &         - \\
CapDec \citeyearpar{capdec}  &      60.1 &      10.2 &      50.2 &       9.3 &      28.7 &      6.0 &      45.9 &       8.3 \\
ViECap \citeyearpar{viecap}  &      61.1 &      10.4 &      64.3 &       9.9 &      65.0 &      8.6 &      66.2 &       9.5 \\
\midrule
\model$^\star$               & \mff{70.1}& \mff{11.2}& \mff{72.5}& \mff{10.9}& \mff{72.1}& \mff{9.6}& \mff{74.0}& \mff{10.5}\\
\bottomrule[2pt]
\end{tabular}
}\caption{Results on the NoCaps validation split. $\star$: without Entity Filtering module in the inference time. $\model$ achieves state of the art in every metric.}\label{tb:nocaps}
\end{table}

\subsection{Text-only Captioning}

We compare our model with other state-of-the-art text-only image captioning models. CapDec~\cite{capdec} and ViECap~\cite{viecap} are based on Clipcap~\cite{clipcap}. They use predefined Gaussian noise for aligning text and image features. Similarly, CLOSE~\cite{close} uses various noise settings, and DeCap~\cite{decap} uses a memory bank. And a recent approach to text-only image captioning, Knight ~\cite{knight} only utilizes text features with a retrieval mechanism, also MeaCap~\cite{meacap} processes retrieved sentences into Subject-Predicate-Object triplets and employs them as additional information. ICSD~\cite{icsd} and SynTIC~\cite{syntic} utilize text-to-image generation models like Stable Diffusion~\cite{stablediffusion} to close the gap. 

\begin{table}[t]
\tabcolsep=2pt
\resizebox{\columnwidth}{!}{
\begin{tabular}{l|cccc|cccc}
\toprule[2pt]
\multirow{2}{*}{\textbf{Method}} & \multicolumn{4}{c|}{\textbf{MSR-VTT}} & \multicolumn{4}{c}{\textbf{MSVD}} \\
& B@4 & M & C & S & B@4 & M & C & S \\
\midrule
ZeroCap \citeyearpar{zerocap} &      2.3  &      12.9 &      5.8  &      -   &      2.9  &      16.3 &      9.6  &      -   \\     
MAGIC \citeyearpar{magic}     &      5.5  &      13.3 &      7.4  &      4.2 &      6.6  &      16.1 &      14.0 &      2.9 \\     
CLMs \citeyearpar{clms}       &      6.2  &      17.8 &      10.1 &      6.5 &      7.0  &      16.4 &      20.0 &      3.1 \\     
CapDec \citeyearpar{capdec}   &      8.9  &      23.7 &      11.5 &      5.9 &      7.9  &      23.3 &      34.5 &      3.2 \\     
EPT \citeyearpar{ept}         &      3.0  &      14.6 &      11.3 &      -   &      3.0  &      17.8 &      17.4 &      -   \\     
Knight \citeyearpar{knight}   &      25.4 & \mff{28.0}&      31.9 & \mff{8.5}&      37.7 & \mff{36.1}&      63.8 &      5.0 \\
\midrule
$\model$                & \mff{27.1}&      25.9 & \mff{38.9}&      6.7 & \mff{40.6}&      34.2 & \mff{83.9}& \mff{6.3}\\
\bottomrule[2pt]
\end{tabular}
}\caption{Results on the Video captioning including MSR-VTT and MSVD. $\model$ achieves state-of-the-art in most metrics.}\vspace{-2mm}\label{tb:videocaptinoing}
\end{table}


\begin{table}[t]
\tabcolsep=3pt
\resizebox{\columnwidth}{!}{
\begin{tabular}{ccc|cccc}
\toprule[2pt]
\textbf{Image-like}    & \textbf{Fusion}  & \textbf{Entity}      & \multicolumn{4}{c}{\textbf{COCO}} \\
\textbf{Retrieval}     & \textbf{Module}  & \textbf{Filtering}   & B@4 & M & C & S          \\
\midrule
$\boldsymbol{\checkmark}$     & $\boldsymbol{\checkmark}$     & $\boldsymbol{\checkmark}$ & \mff{30.8} & \mff{26.7} & \mff{108.0} & \mff{20.3} \\
$\checkmark$     &                  & $\checkmark$     & 29.2 & 26.0 & 104.0 & 19.9 \\
$\checkmark$     & $\checkmark$     &                  & 28.5 & 26.0 & 102.0 & 20.0 \\
                 &                  & $\checkmark$     & 27.2 & 24.7 & 97.3 & 18.5 \\
$\checkmark$     &                  &                  & 27.7 & 25.6 & 99.0 & 19.4 \\
                 &                  &                  & 27.2 & 24.8 & 92.9 & 18.2 \\

\bottomrule[2pt]
\end{tabular}
}\caption{Ablation studies of the key components of $\model$.}\label{tb:ablation_main_component}
\end{table}
\begin{table}[t]
\tabcolsep=2pt
\resizebox{\columnwidth}{!}{
\begin{tabular}{c|ccc|cccc}
\toprule[2pt]
\textbf{Design Choice} & \multirow{2}{*}{\textbf{Pre-$\epsilon$}} & \multirow{2}{*}{\textbf{Post-$\epsilon$}} & \multirow{2}{*}{\textbf{Retrieval}}  & \multicolumn{4}{c}{\textbf{COCO}} \\
\textbf{Reference}     &                                          &                                  &                             & B@4 & M & C & S          \\
\midrule
\multicolumn{1}{l|}{ViECap}          & & &                                        & 27.2 & 24.8 & 92.9 & 18.2 \\
\multicolumn{1}{l|}{Smallcap}        &              &              & $\checkmark$ & 23.5 & 24.2 & 88.5 & 18.2 \\
\multicolumn{1}{l|}{Knight}          &              & $\checkmark$ & $\checkmark$ & 26.0 & 24.6 & 92.9 & 18.3 \\
\multicolumn{1}{l|}{Knight + ILR}    & $\checkmark$ & $\checkmark$ & $\checkmark$ & 27.2 & 25.0 & 93.9 & 18.3 \\
\multicolumn{1}{l|}{\model }         & $\boldsymbol{\checkmark}$ &              & $\boldsymbol{\checkmark}$ & \mff{28.5} & \mff{26.0} & \mff{102.0} & \mff{20.0} \\
\bottomrule[2pt]
\end{tabular}
}\caption{Importance of noise injection timing of \textbf{Image-like Retrieval}. \textbf{Pre-$\epsilon$} refers to noise injection before retrieval, and \textbf{Post-$\epsilon$} refers to noise injection to retrieved features.}\label{tb:ablation_image-like-retrieval}
\end{table}

\begin{table}[t]
\tabcolsep=12pt
\resizebox{\columnwidth}{!}{
\begin{tabular}{c|cccc}
\toprule[2pt]
\textbf{$k$ retrieved}  & \multicolumn{4}{c}{\textbf{COCO}} \\
\textbf{sentences}    & B@4 & M & C & S          \\
\midrule
3   & 28.1 & 25.7 & 100.0 & 19.5 \\
\mff{5}   & \mff{28.5} & \mff{26.0} & \mff{102.0} & \mff{20.0} \\
7   & 28.2 & \mff{26.0} & 101.7 & 19.8 \\
\bottomrule[2pt]
\end{tabular}
}\caption{Ablation studies of the number of retrieved captions $k$ for \textbf{Fusion Module}.}\label{tb:ablation_FM_k_retrieved_sentences}
\end{table}

\begin{table}[ht]
\tabcolsep=3pt
\resizebox{\columnwidth}{!}{
\begin{tabular}{cc|cccc}
\toprule[2pt]
\textbf{Transformer}  & \textbf{Cross-Attention} & \multicolumn{4}{c}{\textbf{COCO}} \\
\textbf{\# Layers}     & \textbf{\# Layers}        & B@4 & M & C & S                   \\
\midrule
\multirow{2}{*}{1}            & 1          &          23.9 &         24.6  &         86.9   &         17.8 \\ 
                              & 4          &          26.2 &         24.4  &         92.8   &         18.0 \\ 
\midrule
\multirow{2}{*}{2}            & 1          &          27.4 &         24.9  &         95.0   &         18.5 \\
                              & 4          &          26.4 &         24.9  &         95.5   &         18.4 \\
\midrule
\multirow{2}{*}{4}            & 1          &          27.4 &         25.5  &         99.7   &         19.1 \\
                              & 4          &          27.9 &         25.8  &         99.1   &         19.4 \\
\midrule
\multirow{2}{*}{\textbf{8}}   & \textbf{1} &          28.3 & \textbf{26.0} & \textbf{102.0} & \textbf{20.0} \\
                              & 4          & \textbf{28.4} &         25.7  &         100.6  &         19.5 \\                 
\bottomrule[2pt]
\end{tabular}
}\caption{Ablation studies of the number of transformer layers and cross-attention layers of the \textbf{Fusion Module}.}\label{tb:ablation_FM_layer_choice}
\end{table}

\begin{table}[ht]
\tabcolsep=3pt
\resizebox{\columnwidth}{!}{
\begin{tabular}{c|cccc|cccc}
\toprule[2pt]
\textbf{$l$ retrieved}  & \multicolumn{4}{c|}{\textbf{COCO}}  & \multicolumn{4}{c}{\textbf{Flickr}} \\
\textbf{sentences}    & B@4 & M & C & S & B@4 & M & C & S          \\
\midrule
5   &          29.9 & 26.4          & 106.1          &          20.2 & \textbf{23.5} &          22.2 &          61.9 & 16.0 \\
7   &          30.3 & 26.5          & 107.2          & \textbf{20.3} & \textbf{23.5} & \textbf{23.0} & \textbf{64.4} & \textbf{17.0}  \\
9   & \textbf{30.8} & \textbf{26.7} & \textbf{108.0} & \textbf{20.3} &          23.4 &          22.6 &          62.9 & 16.6  \\
\bottomrule[2pt]
\end{tabular}
}\caption{Ablation studies of the number of retrieved sentences $l$ for \textbf{Entity Filtering}.}\label{tb:ablation_EF_k_retrieved_sentences}
\end{table}

\begin{table}[ht]
\tabcolsep=3pt
\resizebox{\columnwidth}{!}{
\begin{tabular}{c|cccc|cccc}
\toprule[2pt]
\multirow{2}{*}{$~~~\boldsymbol{\tau}~~~$}  & \multicolumn{4}{c|}{\textbf{COCO}} & \multicolumn{4}{c}{\textbf{Flickr30k}} \\
                             & B@4 & M & C & S & B@4 & M & C & S  \\
\midrule
1   &           6.5 &          18.7 &            6.4 &          17.0 &           6.8 &          18.9 &           3.9 & 15.4 \\
2   &          21.4 &          26.5 &           80.3 &          21.0 &          18.9 & \textbf{23.4} &          52.2 & \textbf{17.9}  \\
3   &          28.1 &          26.8 &          103.6 & \textbf{21.1} &          23.5 &          23.0 & \textbf{64.4} & 17.0  \\
4   &          30.2 & \textbf{26.7} &          107.7 &          20.7 & \textbf{23.8} &          22.3 &          61.1 & 15.9  \\
5   & \textbf{30.8} & \textbf{26.7} & \textbf{108.0} &          20.3 & \textbf{23.8} &          21.9 &          59.1 & 15.3  \\
6   &          30.4 &          26.4 &          106.2 &          19.9 &          23.6 &          21.7 &          57.3 & 15.0  \\
7   &          30.0 &          26.1 &          104.6 &          19.6 &          23.6 &          21.6 &          56.5 & 14.8  \\
8   &          29.8 &          26.0 &          103.4 &          19.4 &          23.7 &          21.6 &          55.9 & 14.7  \\
\bottomrule[2pt]
\end{tabular}
}\caption{Ablation studies of heuristic threshold $\tau$ of \textbf{Entity Filtering}.}\label{tb:ablation_EF_heuristic}
\end{table}

\begin{table}[ht]
\tabcolsep=3pt
\resizebox{\columnwidth}{!}{
\begin{tabular}{l|cccc|cccc}
\toprule[2pt]
\multicolumn{1}{c|}{\multirow{2}{*}{$\boldsymbol{\tau}_\textbf{adap}$}}  & \multicolumn{4}{c|}{\textbf{COCO}} & \multicolumn{4}{c}{\textbf{Flickr30k}} \\
                             & B@4 & M & C & S & B@4 & M & C & S  \\
\midrule
\multicolumn{9}{l}{$\boldsymbol{\mathop{Lognormal}(\mu, \sigma^2)}$} \\
$\mu$             &          22.0 &          26.6 &           83.8 & \textbf{21.1} &          19.0 & \textbf{23.4} &          52.7 & \textbf{17.9}  \\
$\mu + \sigma$    &          29.1 & \textbf{26.7} & \textbf{106.6} &          20.7 &          22.0 &          22.9 & \textbf{63.0} & 17.2  \\
$\mu + 2\sigma$   & \textbf{29.6} &          26.1 &          103.5 &          19.6 & \textbf{23.3} &          21.8 &          58.1 & 15.3  \\
\midrule
\multicolumn{9}{l}{$\boldsymbol{N(\mu, \sigma^2)}$} \\
$\mu$             &          24.9 & \textbf{26.7} &           95.9 & \textbf{21.1} &          19.2 & \textbf{23.2} &          55.6 & \textbf{17.7}  \\
$\mu + \sigma$    & \textbf{30.1} &          26.6 & \textbf{107.5} &          20.4 &          22.3 &          22.5 & \textbf{62.3} & 16.4  \\
$\mu + 2\sigma$   &          29.8 &          26.2 &          104.7 &          19.7 & \textbf{23.4} &          21.9 &          58.5 & 15.5  \\
\midrule
Best (H)          & 30.8 & 26.7 & 108.0 & 20.3 & 23.5 & 23.0 & 64.4 & 17.0  \\
\bottomrule[2pt]
\end{tabular}
}\caption{Ablation studies of adaptive threshold $\tau_\text{adap}$ of \textbf{Entity Filtering.}}\label{tb:ablation_EF_adaptive}
\end{table}

\subsection{In-domain Captioning}

We benchmark our $\model$ on in-domain settings in \Tref{tb:indomain} including COCO and Flickr30k. We compare our methods with previous state-of-the-art in text-only image captioning. Our $\model$ dominates every metric in the COCO dataset compared to models that utilize larger models \cite{close, knight} and have complex training time \cite{icsd, syntic}. Also, in Flickr30k, $\model$ shows decent performance in BLEU@4 and METEOR and achieves the best scores in CIDEr and SPICE.

\subsection{Cross-domain Captioning}

We validate $\model$'s transfer ability through diverse domains, including the NoCaps validation set and cross-domain from COCO $\rightarrow$ Flickr30k and vice versa. In NoCaps, we use the same model trained in the COCO domain to test how the model recognizes unseen objects during training. In the NoCaps validation split, our $\model$ performs the best in every metric and every domain compared to previous state-of-the-art text-only image captioning models \cite{decap, capdec, viecap}. Also, in cross-domain settings between COCO and Flickr30k, $\model$ wins state-of-the-art in most metrics and the second best in some metrics.

\subsection{Video Captioning}

In video captioning, we train our model in the same manner as previous experiments. First, we perform Image-like Retrieval on the corpus from each video captioning dataset MSVD~\cite{msvd} and MSR-VTT~\cite{msrvtt}. For inference time, we sample 5 images from input video and calculate the average of their CLIP image features. We also retrieve 5 sentences from each sampled image, 25 in total, and also calculate the average of CLIP text features per image. Most of the metrics in both datasets, $\model$, fulfills state-of-the-art performance, except METEOR. 

\begin{figure}[t]
    \centering
    \includegraphics[width=\linewidth]{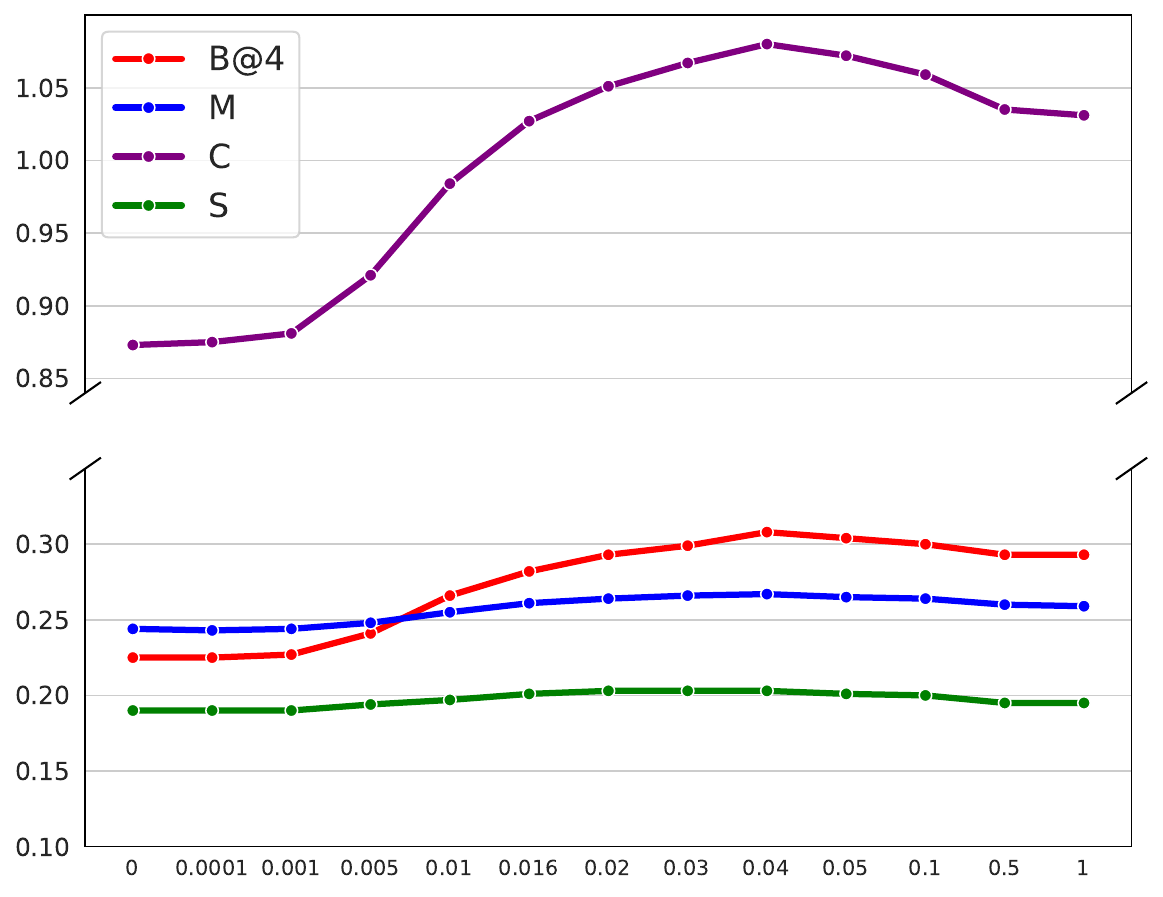}
    \caption{Hyper-parameter search for finding best $\sigma_r$ used in Image-like Retrieval. All experiments are conducted with the COCO test set. The X-axis denotes $\sigma_r^2$, and the Y-axis denotes scores of commonly used captioning metrics BLEU@4 (B@4), METEOR (M), CIDEr (C), and SPICE (S).}
    \label{fig:analysis3}
\end{figure}

\subsection{Ablation Study}

We conduct extensive experiments to identify the impact of each key component in $\model$, Image-like Retrieval (\textbf{ILR}), Fusion Module(\textbf{FM}), and Frequency-based Entity Filtering(\textbf{EF}). Also, for each component, we search the best hyper-parameter in the COCO test split with an in-domain setting. 

\textbf{Key Components:} We check the strength of each component by detaching from our best model, which consists of all 3 components \Tref{tb:ablation_main_component}. First, removing \textbf{FM}, we simply concatenate the input text feature and retrieved features after applying dimension mapping layers $f_{l_1}$ and $f_{l_2}$ and passing it to the caption decoder. Removing \textbf{EF} is simply applying entity extraction via CLIP classifier like \cite{viecap} does. Demounting \textbf{ILR} makes inaccessible to retrieval features solely using input features; hence \textbf{FM} can't exist without \textbf{ILR}. Adding more components into the baseline, we can explicitly notice performance improvement. So, using all three key components constitutes a state-of-the-art model, which is $\model$. 
Note that $\model$ has 2 variants, $\model$ and $\model^\star$, with \textbf{EF} and without \textbf{EF} respectively. To see a full comparison of various datasets, including in-domain, cross-domain, and video captioning, refer to \Tref{tb:ablation_etc}.

\textbf{Image-like Retrieval:} It is crucial to identify adequate timing for injecting noise into text features for successful text-to-text retrieval that imitates image-to-text retrieval. We can split injecting timing into Pre-$\epsilon$ and Post-$\epsilon$. We find that our setting which injects noise before performing retrieval is the best among all possible combinations. We can verify this in \Tref{tb:ablation_image-like-retrieval}. The first column of the table indicates how the model performs retrieval, just for easy understanding of noise injection in retrieval.

\textbf{Fusion Module:} We utilize a cross-attention layer and transformer layer for mapping the network. In \Tref{tb:ablation_FM_layer_choice}, we try multiple combinations of the number of each layer. The more layers we use, the more performance gain we can get until the number of transformer layers is 4. The performance gain is also observed when we use 8 transformer layers but it is so slight. Increasing the number of cross-attention layers is effective when the transformer layer is small, but the tendency does not last while the transformer layer grows. We conclude using 8 transformer layers and a single cross-attention layer shows the best performance. For a fair comparison, we detach the EF module. 
Also, the number of retrieved captions is crucial. We conduct ablation studies to find optimal $k$, which can be found in \Tref{tb:ablation_FM_k_retrieved_sentences}. 

\textbf{Frequency-based Entity Filtering:} We need to choose 1) how many retrieved sentences $l$, to use and 2) the threshold $\tau$, for filtering nouns for \textbf{EF} to extract accurate and diverse entities. The former can be found in  \Tref{tb:ablation_EF_k_retrieved_sentences}, note that in different domains, optimal $l$ may vary. For the COCO domain, using $l$ as 9 shows the best performance, while 7 is the best in Flickr30k. 

We find the best threshold setting in a heuristic and adaptive way. In the former case \Tref{tb:ablation_EF_heuristic}, we set $\tau$ ranging from 1 to 8, which is the minimum and maximum value of the given setting. Above 8, performance freeze due to none of the entities being retrieved. In the COCO test, we use $l = 9$ and $l = 7$ in the Flickr30k test split. We notice that each domain has different optimal $\tau$, COCO at 5 and Flickr30k at 3 for the CIDEr score. In contrast to the heuristic way, we can assume such distribution exists from frequencies $F$. We try Gaussian distribution and Log-normal distribution with $\mu$, $\mu+\sigma$, and $\mu+2\sigma$, capturing upper 50\%, 15.8\%, and 2.2\% based on the frequency of entity. In \Tref{tb:ablation_EF_adaptive}, we observe $\tau_\text{adap}=\mu+\sigma$ almost reproduces the performance of global optimal in the heuristic threshold. If ground truth does not exist or computing resource is limited, the adaptive threshold becomes attractive.

\section{Conclusion}
In this paper, we propose a zero-shot captioning method, $\model$,  through text-only training. $\model$ performs \emph{Image-like Retrieval} to address the gap between image-to-text retrieval and text-to-text retrieval, \emph{Fusion Module} for interaction be-
tween existing and additional representations, and \emph{Frequency-based Entity Filtering} during inference time to extract frequently occurring entities from the retrieved sentences. Our method can be easily applied to various tasks and provides valuable guidance for retrieval-based methods in a text-only setting. It offers clear and precise information to LLMs without relying on a limited vocabulary. The simplicity and robustness of $\model$ are demonstrated through state-of-the-art performance across various datasets in image and video captioning.
The future direction of our method includes the extension of our method on more complex datasets, such as region-based captioning~\cite{kim2019dense,kim2021dense} or visual question answering~\cite{cho2023empirical,cho2023generative}, which suffer from data issues.

\section{Limitations}

We demonstrate that $\model$ exhibits superior performance across various image captioning and video captioning datasets compared to other zero-shot image captioning models with text-only training. However, the optimal value of $\epsilon_r$ for Image-like Retrieval currently requires a heuristic approach to determine. We leave the task of finding a more convenient method for determining the optimal $\epsilon_r$ as future work to further improve image captioning models with text-only training.

\section*{Acknowledgements}
This was partly supported by the Institute of Information \& Communications Technology Planning \& Evaluation (IITP) grant funded by the Korean government(MSIT) (No.RS-2020-II201373, Artificial Intelligence Graduate School Program(Hanyang University)) and the National Research Foundation of Korea(NRF) grant funded by the Korea government(MSIT) (No. RS-2023-00245661).

\bibliography{custom}

\appendix\label{sec:appendix}

\section{Image-like Retrieval}

\begin{table}[ht]
\tabcolsep=7pt
\resizebox{\columnwidth}{!}{
\begin{tabular}{l|cccc}
\toprule[2pt]
\multirow{2}{*}{\textbf{Method}}  & \multicolumn{4}{c}{\textbf{COCO}} \\
    & B@4 & M & C & S          \\
\midrule
Knight   & 27.8 & \textbf{26.4} & 98.9 & 19.6\\
Knight + ILR   & \textbf{29.8} & 25.6 & \textbf{102.7} & \textbf{19.7}\\
\bottomrule[2pt]
\end{tabular}
}\caption{Effect of \textbf{Image-like Retrieval} on Knight.}\label{tb:knight_improvement}
\end{table}

\begin{table}[H]
\tabcolsep=3pt
\resizebox{\columnwidth}{!}{
\begin{tabular}{c|c|c|c|c|c}
\toprule[2pt]
\textbf{HyperParameters}        & \textbf{COCO} & \textbf{Flickr30k} & \textbf{NoCaps} & \textbf{MSVD} & \textbf{MSR-VTT} \\
\midrule
\textbf{Epochs}                          & 5             & 30                 & -               & 10            & 10               \\
$\boldsymbol{l}$ & 9             & 7                  & 7               & 7             & 7                \\
$\boldsymbol{\tau}$           & 5             & 3                  & 3               & 5             & 6                \\
\bottomrule[2pt]
\end{tabular}
}\caption{Hyperparameter table.}\label{tb:hyperparameter}
\end{table}

\label{appendixD}

\begin{table*}[t]
\tabcolsep=5pt
\resizebox{\textwidth}{!}{
\begin{tabular}{l|cc|cc|cc|cc|cc|cc|cc|cc|cc|cc}
\toprule[2pt]
 & \multicolumn{4}{|c}{\textbf{In$-$domain}} & \multicolumn{12}{|c}{\textbf{Cross$-$domain}} & \multicolumn{4}{|c}{\textbf{Video Captioning}} \\
\midrule
\multirow{3}{*}{\textbf{Method}} & \multicolumn{2}{c|}{\multirow{2}{*}{\textbf{COCO}}} & \multicolumn{2}{c|}{\multirow{2}{*}{\textbf{Flickr}
}} & \multicolumn{8}{c|}{\textbf{COCO $\Longrightarrow$ NoCaps Val}} & \multicolumn{2}{c|}{\multirow{2}{*}{\textbf{COCO $\Longrightarrow$ Flickr}}} & \multicolumn{2}{c|}{\multirow{2}{*}{\textbf{Flickr $\Longrightarrow$ COCO}}} & \multicolumn{2}{c|}{\multirow{2}{*}{\textbf{MSR-VTT}}} & \multicolumn{2}{c}{\multirow{2}{*}{\textbf{MSVD}}} \\
                        & &  & &  & \multicolumn{2}{c|}{In} & \multicolumn{2}{c|}{Near} & \multicolumn{2}{c|}{Out} & \multicolumn{2}{c|}{Entire} & &  & &  & &  & & \\
                        & C & S & C & S & C & S & C & S & C & S & C & S & C & S & C & S & C & S & C & S \\
\midrule
ViECap          & 92.9 & 18.2  & 47.9 & 13.6 & 61.1 & 10.4 & 64.3 & 9.9  & 65.0 & 8.6 & 66.2 & 9.5  & 38.4 & 11.2 & 54.2 & 12.5 & -    & -   & -    & -   \\
Knight          & 98.9 & 19.6  & 56.3 & 16.3 & -    & -    & -    & -    & -    & -   & -    & -    & 48.9 & 14.2 & 64.4 & 15.1 & 31.9 & \mff{8.5} & 63.8 & 5.0 \\
\midrule
$\model^\star$  & 102.0 & 20.0 & 59.8 & 15.8 & 70.1 & 11.2 & \mff{72.5} & 10.9 & \mff{72.1} & \mff{9.6} & \mff{74.0} & 10.5 & 47.5 & 12.7 & 60.7 & 13.6 & 20.8 & 4.1 & 40.2 & 3.4 \\
$\model$        & \mff{108.0} & \mff{20.3} & \mff{64.4} & \mff{17.0} & \mff{75.8} & \mff{12.4} & 72.3 & \mff{11.6} & 60.2 & 8.9 & 70.5 & \mff{10.8} & \mff{59.2} & \mff{15.6} & \mff{76.3} & \mff{17.3} & \mff{38.9} & 6.7 & \mff{83.9} & \mff{6.3} \\
\bottomrule[2pt]
\end{tabular}
}\caption{Overall comparison among baselines and $\model$. $\star$: without Entity Filtering module in the inference time. }\label{tb:ablation_etc}
\end{table*}

We observe that Image-like Retrieval is also applicable to other models that employ text-to-text retrieval~\cite{knight}. Based on \Fref{fig:analysis3}, we perform \textbf{ILR} with $\epsilon_r=0.04$ in the training time of Knight. In the COCO test set, every metric except METEOR is improved compared to vanilla Knight~\cite{knight}, verifying the effectiveness of our \textbf{ILR}.

\section{Hyperparameter}

We include the details about our experiments in each dataset in \Tref{tb:hyperparameter}.

\section{Comparison with Baselines}

We compare baselines \cite{viecap, knight} with $\model$ and $\model^\star$ in every domain, including in-domain captioning, cross-domain captioning, and video captioning. Results can be found in \Tref{tb:ablation_etc}.

\section{Qualitative Results}

\begin{figure*}[ht]
    \centering    \includegraphics[width=0.9\textwidth]{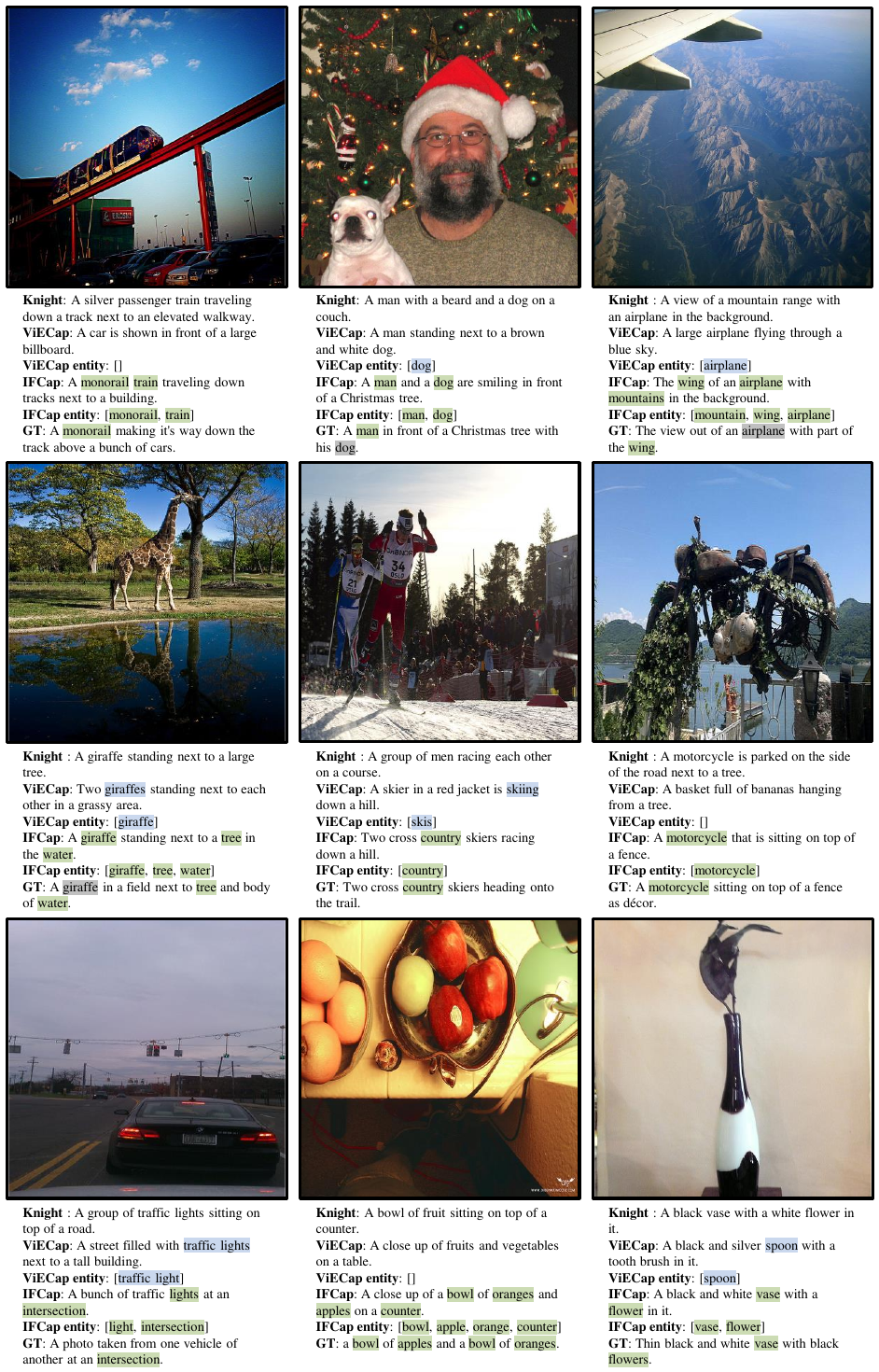}
    \caption{Qualitative result on the COCO test set. We highlight the retrieved entities and their appearance in the generated captions with $\colorbox{rgb:red!30,0.41;green!30,0.52;blue!30,0.24}{\model}$, $\colorbox{rgb:red!30,0.26;green!30,0.53;blue!30,0.96}{ViECap}$ and $\colorbox{rgb:red!30,0.5;green!30,0.5;blue!30,0.5}{Intersection}$.}
    \label{fig:qualitative_result}
\end{figure*}

We show additional qualitative results in \Fref{fig:qualitative_result}.

\end{document}